\documentclass[10pt,twocolumn,letterpaper]{article}

\usepackage{wacv}
\usepackage{times}
\usepackage{epsfig}
\usepackage{graphicx}
\usepackage{amsmath}
\usepackage{amssymb}
\usepackage{multirow}
\usepackage{booktabs}
\usepackage{bbm}
\usepackage{adjustbox}
\usepackage{array}
\usepackage{siunitx}
\usepackage{graphicx}
\usepackage{capt-of}
\usepackage{mathtools}
\usepackage{stmaryrd}

\def\wacvPaperID{1242} %
\wacvfinalcopy %
\ifwacvfinal
\fi

\ifwacvfinal
\usepackage[pagebackref=true,breaklinks=true,colorlinks,bookmarks=false]{hyperref}
\else
\usepackage[pagebackref=true,breaklinks=true,colorlinks,bookmarks=false]{hyperref}
\fi

\ifwacvfinal
\else
\fi

\begin{document}

\title{Interpretable and Trustworthy Deepfake Detection via Dynamic Prototypes}
 \vspace{-0.2cm}
\author{Loc Trinh, Michael Tsang, Sirisha Rambhatla, Yan Liu\\
University of Southern California\\
Los Angeles, CA 90089\\
{\tt\small \{loctrinh, tsangm, sirishar, yanliu.cs\}@usc.edu}\\
\vspace{-0.8cm}
}

\maketitle
\newcommand{\proposed}{{\tt DPNet}}

\begin{abstract}
In this paper we propose a novel human-centered approach for detecting forgery in face images, using dynamic prototypes as a form of visual explanations. Currently, most state-of-the-art deepfake detections are based on black-box models that process videos frame-by-frame for inference, and few closely examine their temporal inconsistencies. However, the existence of such temporal artifacts within deepfake videos is key in detecting and explaining deepfakes to a supervising human. To this end, we propose Dynamic Prototype Network (\proposed) -- an interpretable and effective solution that utilizes dynamic representations (i.e., \textit{prototypes}) to explain deepfake temporal artifacts. Extensive experimental results show that {\proposed} achieves competitive predictive performance, even on unseen testing datasets such as Google's DeepFakeDetection, DeeperForensics, and Celeb-DF, while providing easy referential explanations of deepfake dynamics.  On top of {\proposed}'s prototypical framework, we further formulate temporal logic specifications based on these dynamics to check our model's compliance to desired temporal behaviors, hence providing trustworthiness for such critical detection systems.
\end{abstract}

\vspace{-0.2cm}
\section{Introduction}
While artificial intelligence (AI) plays a major role in revolutionizing many industries, it has also been used to generate and spread malicious misinformation.  In this context, \textit{Deepfake} videos -- which can be utilized to alter the identity of a person in a video -- have emerged as perhaps the most sinister form of misinformation, posing a significant threat to communities around the world \cite{Vaccari, Ingram, Toews, Turton}, especially with election interference or nonconsensual fake pornography \cite{Cahlan, Hao}. Therefore, as deepfakes become more pervasive, it is critical that there exists algorithms that can ascertaining the trustworthiness of online videos. 

\begin{figure*}[ht]
\centering
\label{fig:one}
\includegraphics[keepaspectratio=false,height=4.5cm, width=.98\textwidth]{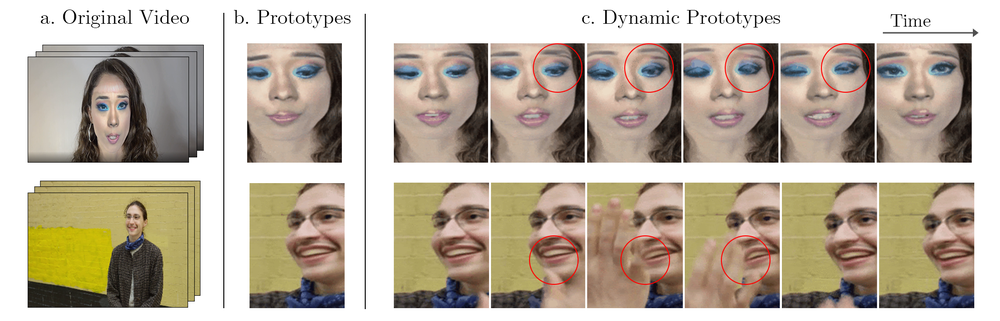}
\caption{\textbf{Examples of static vs. dynamic explanations for deepfake videos.} Qualitatively, seeing temporal artifacts allow a human to quickly determine whether a video is real or fake. Red circles indicate regions of interest. Best view as GIFs (See Appendix \ref{apd:gifs}).}
\vspace{-0.4cm}
\end{figure*}

To address this challenge, a series of excellent works has been conducted on detecting deepfakes \cite{DFsurvey, Matern, Afchar, Rossler, li2020face}. While these work have achieved good progress towards the prediction task to a certain extent, there is still significant room for improvement. First, even though existing work focus on the detection problem, very few of them address the interpretability and trustworthiness aspects. Currently, most existing solutions draw bounding boxes around a face and label it with \textit{fakeness} probabilities. Rather, it might be more fruitful to explain \textit{why} a model predicts a certain face as real or fake, such as which parts of the face the model believes are forged, and where is it looking to yield this prediction. This is crucial for a human to understand and trust the content verification systems. Second, it is known that humans can instantaneously detect deepfake videos after observing certain unnatural dynamics, due to the distortions induced by deepfake generative models, which are generally harder to hide \cite{Korshunov2018DeepFakesAN, Petrov2020DeepFaceLabAS, Zhu2017UnpairedIT}. This too would also be a viable explanation for a system to return as humans can quickly see and understand abnormal movements (Figure \ref{fig:one}). Yet most state-of-the-art deepfake detection techniques only analyze a potential video frame-by-frame, and few have explored these temporal inconsistencies \cite{Sabir, qian2020thinking, masi2020two}. As a result, there is a need for an interpretable deepfake detection method that \textit{both} considers temporal dynamics and at the same time provides human-accessible explanations and insights into the inconsistencies within deepfake videos.

To this end, we propose {\proposed} -- an interpretable prototype-based neural network that captures dynamic features, such as unnatural movements and temporal artifacts, and leverages them to explain \textit{why} a particular prediction was made. Specifically, {\proposed} works by first learning the prototypical representations of the temporal inconsistencies within the latent space, by grouping the patch-wise representation of real video closer together while pushing those of fake videos farther away. Then, it makes predictions based on the similarities between the dynamics of a test video and a small set of learned dynamic prototypes. Lastly, the prototypes are then intermittently projected to the closest representative video patch from the training dataset, which yields an immediate human-understandable interpretation of the learned dynamic prototypes. 

The primary advantages of {\proposed} are as follows: \vspace{-0.2cm}
\begin{itemize}
\item \textbf{Faithful explanations via case-based reasoning}: {\proposed} follows a case-based reasoning approach that utilizes previously learned dynamics - as a piece of evidence (i.e \textit{cases}) - to tackle an unseen testing video. This also allows the model to explain \textit{why} a certain prediction was made, in a way that is reflective of the network's underlying computational process. \vspace{-0.2cm}
\item \textbf{Visual dynamic explanations}: {\proposed} provides explanations in the form of visual dynamics (video clips) via the learned dynamic prototypes, each of which points to a temporal artifact that is accessible and easy for humans to understand.
\vspace{-0.2cm}
\item \textbf{Temporal logic specifications}: Lastly, the dynamic prototypes learned by the network can additionally be used to formulate temporal logic specifications. This allows auditors to check the robustness of the model and verify whether certain temporal behaviors are obeyed throughout the lengths of the videos.
\end{itemize}

\section{Related Work}
\subsection{Face forgery detection} \vspace{-0.1cm} Early face forensic work focus on  hand-crafting facial features, such as eye color and missing reflections \cite{Matern}, 3D head poses \cite{Yang}, and facial movements \cite{Agarwal,OpenFace2}.  However, these approaches do not scale well to larger and more sophisticated deepfakes. To address this problem, researchers leverage recent advances in deep learning to automatically extract discriminative features for forgery detection \cite{Rossler, Nguyen1, Nguyen2, wang2020fakespotter}. Previous work achieved state-of-the-arts by fine-tuning ImageNet-based model, such as Xception \cite{Rossler}. Other work examine spatial pyramid pooling module to detect resolution-inconsistent facial artifacts DSP-FWA \cite{Li}, low-level features and convolutional artifacts \cite{Afchar, chai2020makes}, or blending artifacts via Face X-ray \cite{li2020face}.  FakeSpotter \cite{wang2020fakespotter} uses layer-wise neuron behaviors as features instead of final-layer neuron output to train a classifier.

Most forgery detection methods process deepfake videos frame-by-frame, and few explore multi-modal and temporal dynamics \cite{Zhou, Amerini}. Recent work using multi-frame inputs \cite{Sabir, masi2020two} and video architecture \cite{qian2020thinking} have shown the competitive potential of leveraging temporal information. Our approach builds on this and examines temporal artifacts both to predict and explain deepfakes to human decision makers.

With more advanced deepfake creations, recent works \cite{cozzolino2018forensictransfer, 8553251, du2019towards} have shown that the performance of current methods \textit{drops} drastically on new types of facial manipulations. In particular, ForensicTransfer \cite{cozzolino2018forensictransfer} proposes an autoencoder-based neural network to transfer knowledge between different but related manipulations. Face X-ray \cite{li2020face} created a blending dataset to help networks generalize across various manipulations, and \cite{masi2020two} creates a novel loss to reshape the latent space, pulling real representations closer and repelling fake representations farther away, both of which have demonstrated valuable generalizability.

\subsection{Interpretable neural networks} \vspace{-0.1cm} One prominent approach to explaining deep neural networks is \textit{posthoc} analysis via gradient \cite{Simonyan, Selvaraju, Sundararajan, bach, Shrikumar} and perturbation-based methods \cite{fong2017interpretable, sato2018interpretable,  Zeiler, Montavon}; however, it is known that these methods do not modify the complex underlying architecture of the network. Instead, another line of research tries to build networks that are interpretable by design, with a \textit{built-in} way to self-explain \cite{Alvarez, Oscar}. The advantage of this approach is that interpretability is represented as \textit{units} of explanation - general concepts and not necessarily raw inputs. This can be seen in the work of Alvarez et. al \cite{Alvarez} for basis concept learning and Kim et. al \cite{kim2014bayesian, Ming} for case-based reasoning and prototype learning. Recently, Chen et al. \cite{Chen} proposed learning prototypes for fine-grained image classification to make predictions based on similarity to class-specific image patches via ProtoPNet. 

On the other hand, although complex deep learning-based video classification models have been developed for video understanding, such as 3D CNN, TSN, TRN, and more \cite{zhoutrn, lin, zhoutrn, Wang, Feichtenhofer}, there is much to be desired in terms of interpretability, especially when compared to intrinsically interpretable models. In contrast, our proposed approach extends fine-grain classification \cite{Chen} and captures fake temporal artifacts as dynamic prototypes, which can be directly visualize to explain predictions to a human being, which is crucial important for face forgery detection.

\begin{figure*}[t]
\centering
\includegraphics[keepaspectratio=false,height=4.5cm, width=.98\textwidth]{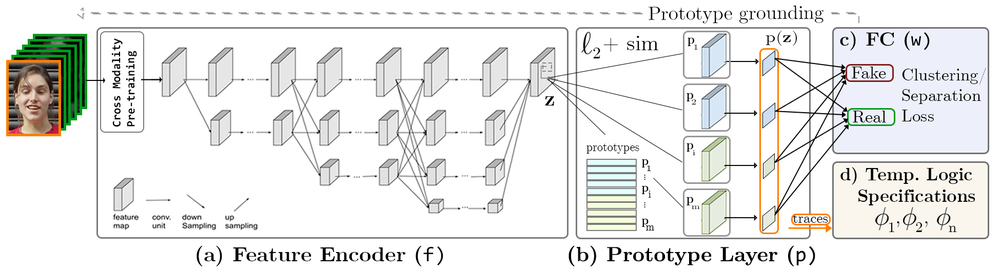}
\caption{\textbf{DPNet video-based face manipulation detection architecture.} Spatial and temporal information are processed via the HRNet feature encoding backbone. The networks learn $m$ prototypes, each is used to
represent some prototypical activation pattern in a patch of the convolutional feature maps, which in turn corresponds to some prototypical dynamic patch in the original spatial/temporal space. Traces from videos along with prototypical activations are then verified via the temporal logic module post-training.}
\label{figure:two}
\vspace{-0.3cm}
\end{figure*}

\subsection{Safety verification} \vspace{-0.1cm} The importance of safety verification, especially in critical domains such as healthcare or autonomous driving, was highlighted when the discovery of adversarial attacks \cite{Szegedy, Goodfellow} prompted many lines of work in robust verification in ML \cite{Lcuyer2018CertifiedRT, Katz2017ReluplexAE, Gowal2018OnTE, Huang2017SafetyVO, wang2018efficient, Carlini2017TowardsET, Madry2017TowardsDL}. To further reason about safety and robustness in time, temporal logic has been broadly in previous work \cite{BOUYER2009323, Padon, sistla1994safety, Dwyer}, and recent work in using temporal logic to verify time-series and NN-based perception system have shown promises \cite{vazquez2018time, Dokhanchi, pmlr-v97-weng19a}. Moreover, the difficulty in building neuro-symbolic system is the question of how to integrate the output representations of deep networks in a propositional form \cite{Besold2017NeuralSymbolicLA}. In contrast, the dynamic prototypes from our interpretable models provide a convenient vehicle for us to specify atomic propositions and formulate temporal logic specifications for videos. This allows users verify the model's compliance to desired temporal behaviors and establish trust within critical detection systems when deployed in a real-world scenario.

\section{Dynamic Prototype Network (DPNet)}
In this section, we introduce our Dynamical Prototype Network ({\proposed}), the loss function, and the training procedure, and the logic syntax. In addition, we highlight the steps that our network took to predict a new video, and how those steps can be interpreted in a human-friendly way.

\subsection{{\proposed}  architecture } \vspace{-0.1cm}
The proposed architecture is shown in  Figure \ref{figure:two}. As shown, {\proposed} consists of four components: the feature encoder $f$, the prototype layer $p$, the fully-connected layer $w$, and the temporal logic verifier. Formally, let $\mathcal{V} = \{(\mathbf{v}_i,y_i)\}_{i=1}^{N}$ be the video dataset, where $\mathbf{v} _i$ be a deepfake video sequence of length $T_{\mathbf{v}_i}$, and $y_i \in \{0,1\}$ is the label for fake/real. Clips are sampled as inputs to the network.

\noindent\textbf{Feature encoder $f(\cdot)$}: The feature encoder  $f$ encodes a processed video input $\textbf{x}_i \in \mathbb{R}^{256 \times 256 \times S}$ into a hidden representation $\textbf{z} \in \mathbb{R}^{H \times W \times C }$. We used HRNet here as the backbone encoder, initialized with ImageNet pretrained weights. The input $\textbf{x}_i $ to the encoder $f$ is formed by stacking one RGB frame with precomputed optical flow fields between several consecutive frames, yielding $S$ channels (Figure \ref{figure:two}a). We let the input to the {\proposed} be a fixed-length $T < T_{\mathbf{v}_i}$, and randomly selected the initial starting frame for $\textbf{x}_i $. This allows us to explicitly describe the motion of facial features between video frames, while \textit{simultaneously} presenting the RGB pixel information to the network. Furthermore, since the optical flow fields can also be viewed as image channels, they are well suited for image-based convolutional networks. The feature encoder $f$ outputs a convolutional tensor $\textbf{z} = f(\textbf{x}_i)$ that is forwarded to the prototype layer.

\noindent\textbf{Prototype layer $p(\cdot)$}: The network learns $m$ prototype vectors $\mathbf{p}_1, \mathbf{p}_2, \dots, \mathbf{p}_m$ of shape $(1,1,C)$ in the latent space, each will be used to represent some dynamical prototypical activation pattern within the convolutional feature maps. The prototype layer $p$ computes the squared $\ell_2$ distance between each the prototype vectors $\mathbf{p}_j$ and each spatial/temporal $patch$ (of shape $(1,1,C)$) within the input feature maps  $\mathbf{z}$. It then inverts the distance to score their similarity. This generates  $m$ similarity maps for each dynamic prototype, and max-pooling yields the most similar activation patch. The shape of the prototype vectors is chosen to represent the smallest facial dynamic patch within $\textbf{z}$. Further patch projection in section \ref{sss:grounding} attributes each prototype to a corresponding prototypical video patch in the original pixel/flow space. Formally, in Figure \ref{figure:two}b, the prototype layer $p$ computes $m$ similarity scores:
\begin{equation}
    p(\mathbf{z}) = [\,p_1(\mathbf{z}),\,\,p_2(\mathbf{z}),\,\,\dots,\,\, p_m(\mathbf{z})\,] ^\top
\end{equation}
where, the similarity score between a prototype $\mathbf{p}_j$ and $\mathbf{z}$, denoted as $p_j(\cdot)$ is given by 
\begin{equation}
    p_j(\mathbf{z}) = \max_{\mathbf{z}' \in \text{patches}(\mathbf{z})} \frac{1}{1 + \|\mathbf{z}' - \mathbf{p}_j\|_2^2}
\end{equation}

\noindent\textbf{Fully-connected layer $w(\cdot)$}: This layer computes weighted sums of similarity scores, $\mathbf{a} = \mathbf{W} ~p(\mathbf{z})$, where $\mathbf{W} \in \mathbb{R}^{K \times m}$ are the weights, and $K$ denotes the number of classes ($K = 2$ for \proposed). We then use a softmax layer to compute the predicted probability as follows, $\hat{y}_i = \frac{\text{exp}(a_i)}{\sum_{j=1}^K \text{exp}(a_j)}.$ Note that, we allocate $m_k$ prototypes for each class $k \in \{0,1\}$ s.t. $\sum_k m_k = m$. In other words, every class is represented by $m_k$ prototypes in the final model.

\subsection{Learning objective} \vspace{-0.1cm}
We aim to learn meaningful forgery representation which ensures that the dynamic prototype vectors a) are close to the input video patches (\textbf{Fidelity}), b) between fake and real artifacts are well-separated (\textbf{Separability}), and c) are interpretable by humans (\textbf{Grounded}). We adopt two widely used loss functions in prototype learning to enforce fidelity and separability. Furthermore, we also introduce a \emph{diversity} loss term to ensure that intra-class prototypes are non-overlapping. We jointly optimize the feature encoder $f$ along with the prototype vectors $\textbf{p}_1, \textbf{p}_2, \dots, \textbf{p}_m$ in the prototype layer $p$ to minimize the the cross-entropy loss on training set, while also regularizing for the desiderata. \vspace{-0.2cm}

\subsubsection{Loss function} \label{sec:loss}
Let $\mathcal{D} = \{(\mathbf{x}_i,y_i)\}_{i=1}^{N}$ be our training dataset, where $\mathbf{x}_i$ is our stacked input extracted from video $\mathbf{v}_i$.

For hyperparameters $\lambda_c, \lambda_s, \text{and\,} \lambda_d$ the overall objective function that we wish to minimize is given by:
\begin{equation}
\begin{aligned}
    \mathcal{L}(\mathcal{D}; \theta) &= CE(\mathcal{D}; \theta) + \lambda_c R_{clus}(\mathcal{D};  \theta) \\ &+ \lambda_s R_{sep}(\mathcal{D}; \theta) + \lambda_d R_{div}(\theta)
\end{aligned} 
\end{equation}

\noindent where $CE(\cdot)$, $R_{clus}(\cdot)$,  $R_{sep}(\cdot)$, $R_{div}(\cdot)$ are the cross-entropy, clustering, separation, and diversity loss, respectively. Here, $\theta$ are the trainable parameters for the feature encoder $f$ and the prototype layer $p$. The cross-entropy loss here imposes prediction accuracy and is given by \vspace{-0.1cm}
\begin{align}
CE( \mathcal{D}; \theta) = \frac{1}{N}\sum_{i=1}^{N} \sum_{k=1}^{K} - \mathbbm{1}\big[y_i = k\big]\log(\hat{y}_k)
\end{align}

\noindent The clustering loss $R_{clus}$ minimizes the squared $\ell_2$ distance between some latent patch within a training video and its closest prototype vector from that class, and is given by \vspace{-0.1cm}
\begin{align}
R_{clus}(\cdot) = \frac{1}{N}\sum_{i=1}^{N} \min_{\mathbf{p}_j\in \textbf{P}_{y_i}} \min_{\mathbf{z} \in \text{patches}({\mathbf{x}_i})} \|\mathbf{z} - \mathbf{p}_j\|_2^2
\end{align} 

\noindent where $\mathbf{P}_{y_i}$ is the set of prototype vectors allocated to the class $y_i$. The separation loss $R_{sep}$ encourages every patch of a \textit{manipulated} training video to stay away from the \textit{real} dynamic prototypes (vice versa), and is given by \vspace{-0.1cm}
\begin{align}
R_{sep}(\cdot) = -\frac{1}{N}\sum_{i=1}^{N} \min_{\mathbf{p}_j \not\in \mathbf{P}_{y_i}} \min_{\mathbf{z} \in \text{patches}({\mathbf{x}_i})} \|\mathbf{z} - \mathbf{p}_j\|_2^2
\end{align} 

\noindent These loss functions have also been commonly used in previous prototypical-based frameworks \cite{Chen, Ming}.

\noindent \textbf{Diversity loss}. Due to intra-class prototypes overlapping, we  propose a cosine similarity-based regularization term which penalizes prototype vectors of the same class for overlapping with each other, given by \vspace{-0.2cm}
\begin{equation}
    R_{div}(\cdot) = \sum_{k=1}^K\sum_{\substack{i \neq j \\ \mathbf{p}_i,\mathbf{p}_j \in {\mathbf{P}_k}}} \max(0, \cos(\mathbf{p}_i, \mathbf{p}_j) - s_{max}\big)
    \label{eq:div}
\end{equation}
where $s_{max}$ is a hyperparameter for the maximum similarity allowed. This cosine similarity-based loss considers the angle between the prototype vectors regardless of their length. It allows us to penalize the similarity between the prototypes up to a threshold, leading to more diverse and expressive dynamic representations. \vspace{-0.2cm}

\subsubsection{Prototype projection and grounding} \label{sss:grounding}
To achieve grounding, while training we intersperse the following projection step after every few epochs. Specifically, we project the prototype vectors to actual video patches from training videos that contain those dynamics as follows,
\begin{equation}
    \textbf{p}_j \leftarrow \text{argmin}_{\textbf{z}' \in \text{patches}(f(\textbf{x}_i))} \|\mathbf{z}' - \mathbf{p}_j\|_2^2 \,\,\,\forall i\,\,\text{s.t.}\,\,y_i = k
\end{equation} 
for all prototype vectors of class $k$, i.e. $\textbf{p}_j \in \textbf{P}_k$. This step projects each prototype vector of a given class to the closest latent representation of a manipulated / genuine video patch that is also from that same class. As a result, the prediction of a test video is made based on the similarities it has with the learned dynamic prototypes. Consequently, the test predictions are \textit{grounded} on the training videos. \vspace{0.1cm}

\subsection{Temporal logic verification via prototypes} \vspace{-0.1cm} \label{ss:temporal}
Our {\proposed} architecture allows for the usage of formal methods to verify the compliance of our model. Given the direct computational path from $f$ to $p$ to $w$, we can check whether the learned prototypes satisfy some desired temporal behaviors. Here, we used Timed Quality Temporal Logic (TQTL) similar to \cite{Dokhanchi}, but instead, we consider each video as a \textit{data stream} with each frame as a \textit{time-step}. We hereby give a brief summary of the TQTL language, and the specifications we used to verify our model.

The set of \textit{Timed Quality Temporal Logic} (TQTL) formulas $\phi$ over a finite set of Boolean-value predicates $Q$ over attributes of prototypes, a finite set of time variables ($V_t$), and a finite set of prototype
indexes ($V_{p}$) is inductively defined according to the following grammar:
\begin{equation}
    \begin{aligned}
        \phi ::=\,&\texttt{true}\,|\,\pi\,|\,\neg\phi\,| \,\phi_1\lor \phi_2\,|\,\phi_1\,\textbf{U}\,\phi_2\,|\\ 
        &\,x \leq y + n\,|\,x.\phi\,|\,\exists p_i@x, \phi\,
    \end{aligned}
\end{equation}
where $\pi \in Q,$ and $\phi_1$ and $\phi_2$ are valid TQTL formulas. $\pi$ has the functional form $\pi \equiv F_\pi(t_{1\dots n}, p_{1\dots m}) \sim C$, where $\sim$ is a comparison operator, i.e. $\sim\,\in \{<, \leq, >, \geq, =, \not=\}$, and $C \in \mathbb{R}$. For example, the predicate for ``the similarity of prototype $\textbf{p}_1$ to an input at time step $2$ is greater than 0.9" is $F(t_2, p_1) > 0.9$. We hereby use $S(\cdot)$ to denote the prototype similarity score.

In the grammar above, $x,y \in V_t, n \in \mathbb{N}, p_i \in V_{p} $, and $\textbf{U}$ is the “until" operator. The time constraints of TQTL are represented in the form of $x \leq y + n$. The freeze time quantifier $x.\phi$ assigns the current time to a variable $x$ before processing the subformula $\phi$. The quantifiers $\exists$ and
$\forall$ respectively existentially and universally quantify over the
prototypes in a given frame. In addition, we use three additional operators: ($\psi$ Implies $\phi$) $\psi \rightarrow \phi \equiv \neg\psi\lor\phi$, (Eventually $\psi$) $\Diamond \psi \equiv \texttt{true}\,\textbf{U}\,\psi$, and (Always $\psi$) $\square\psi\equiv \neg\Diamond\neg\psi$. The semantics of TQTL can be find in the Appendix \ref{apd:TQTL}.

From the dynamic prototypes, we define \textit{specifications} to check the robustness of our model. We verify that if our model predicts $fake$ for a testing video $V$, throughout the video, there exists a clip starting at time $t$ where a prototype $\textbf{p}_i \in \textbf{P}_{fake}$ is most similar to it compared to all prototypes of the $real$ class $\textbf{p}_k \in \textbf{P}_{real}$ for all time $t'$ s.t. $0 \leq t' \leq T_{V}$. This verifies that there exists of a key region of the video that our prototypes `see' $fake$ strongly. The formula $\phi_{1}$ denotes this \textit{key-frame} specification, $\mathcal{F}$ = fake, $\mathcal{R}$ = real:
\begin{align*}
    \phi_{1} =\,\,\,&\Diamond(t.\exists p_k @t, Class(V)=\mathcal{F}(\mathcal{R}) \wedge p_k \in P_{\mathcal{F}(\mathcal{R})}\\
    &\rightarrow \square (t'. ((0 < t' \wedge t' < T_V) \\
    &\rightarrow \forall p_j@t', p_j \in P_{\mathcal{R}(\mathcal{F})} \wedge S(t, p_k) > S(t', p_j)\,\,)))
\end{align*}
Moreover, we can specify that if a prototype is \textit{non-relevant}, its similarity to a frame should be consistently low throughout. An example of this safety specification is that a prototype representing a $fake$ temporal artifact should not be highly activated in a $real$ video at any time. We specify this notion below, where numerical values are \textit{user-specified thresholds} that control the strictness of the specification. The formula $\phi_{2}$ denotes this \textit{non-relevance} specification:
\begin{align*}
    \phi_{2} =\,\,\,&\square(t.\forall p_i@t,  Class(V) = \mathcal{F}(\mathcal{R}) \wedge p_i \in P_{\mathcal{R}(\mathcal{F})}\\
    &\rightarrow S(t,p_i) < 0.4 \wedge \square(t'.(t \leq t' \wedge t' \leq t + 5) \\
    &\rightarrow |S(t',p_i) - S(t,p_i)| < 0.1 ))
\end{align*}

\section{Experiments}

\begin{figure*}[t] \label{fig:three}
\centering
\includegraphics[keepaspectratio=false,height=8.9cm, width=.95\textwidth]{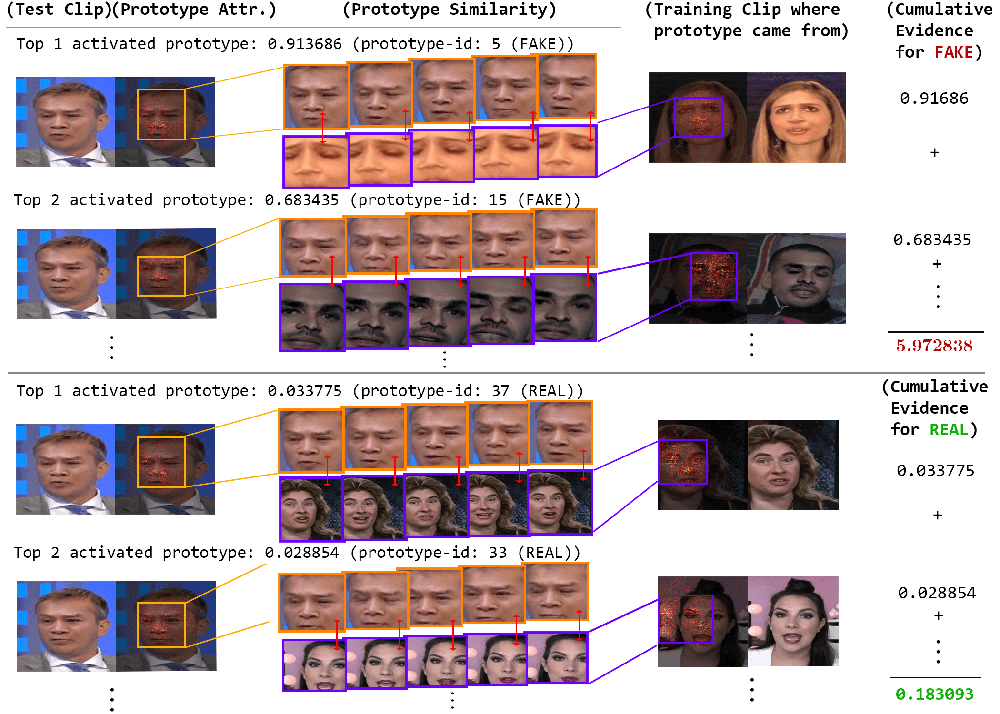}
\caption{\textbf{The reasoning process of the DPNet.} The prediction for video clip is based on the similarity comparison between the latent input representations against the learned prototypes. The network tries to find evidence for manipulated/genuine by looking at which spatial/temporal patch that was mostly activated by the dynamic prototypes. Similarity scores between class-specific prototypes are summed.} \vspace{-0.4cm}
\end{figure*}

\subsection{Experimental settings} \vspace{-0.1cm}
\noindent\textbf{Training datasets.} For training our networks, we use the FaceForensics++ (FF++) \cite{Rossler} dataset, which consisted of 1000 original video sequences that have been manipulated with four face manipulation methods: DeepFakes (DF) \cite{deepfakes}, Face2Face (F2F) \cite{10.1145/3292039}, FaceSwap (FS) \cite{faceswap}, and NeuralTextures (NT) \cite{thies2019deferred}. FF++ provides ground truth manipulation masks showing which part of the face was manipulated. We preprocess the videos by dumping the frames and crops out facial areas based on facial landmarks \cite{mtcnn}.

{\renewcommand{\arraystretch}{1.1}
\begin{table}[t]
\centering
\caption{\textbf{Basic information of training and testing datasets.}
\cite{Li2019CelebDFAL}}
\begin{adjustbox}{width=.97\columnwidth}
\begin{tabular}{l|cc|cc}
\hline
\multirow{2}{*}{Dataset} & \multicolumn{2}{c|}{Real} & \multicolumn{2}{c}{Fake} \\  \cline{2-3}\cline{4-5}
& Video        & Frame        & Video         & Frame \\\hline
FaceForensics++  (FF++) \cite{Rossler} & 1000 & 509.9k & 4000 & 2039.6k \\ \cline{2-5}
DeepFakeDetection (DFD) \cite{DDD_GoogleJigSaw2019} & 363 & 315.4k & 3068 & 2242.7k \\ \cline{2-5}
DeeperForensics-1.0 \cite{jiang2020deeperforensics10} & 0 & 0.0k & 11000 & 5608.9k \\ \cline{2-5}
Celeb-DF \cite{li2020celeb} & 590 & 225.4k & 5639 & 2116.8k \\
\hline
\end{tabular}
\end{adjustbox}
\label{table:one}
\vspace{-0.6cm}
\end{table}}

\noindent\textbf{Testing datasets.} To evaluate the cross-dataset generalizability of our aprpoach, we use the following datasets: 1) FaceForensics++ \cite{Rossler} (FF++); 2) DeepfakeDetection (DFD) \cite{DDD_GoogleJigSaw2019} including 363 real and 3068 deepfake videos released by Google in order to support developing deepfake detection methods; 3) DeeperForensics-1.0 \cite{jiang2020deeperforensics10} - a large scale dataset consisting of 11, 000 deepfake videos generated with high-quality collected data vary in identities, poses, expressions, emotions, lighting conditions, and 3DMM blendshapes; 4) Celeb-DF \cite{li2020celeb}, a new DeepFake dataset of celebrities with 408 real videos and 795 synthesized video with reduced visual artifacts. (Table~\ref{table:one}).

\vspace{0.1cm}
\noindent\textbf{Implementation details.}
During {\proposed} training phases, the input is formed by stacking $1$ RGB frame followed by $10$ pre-computed optical flow fields that are uniformly separated. {\proposed} uses a pre-trained HRNet as a backbone network, warm started with ImageNet pre-trained weights. Since our input contains temporal frames, we also perform cross-modality pre-training to average and increase the number of channels in the first conv. layer.  The encoder $f$ and prototypes $\textbf{p}$ were trained with learning rates of  $2e^{-4}$ and  $1e^{-3}$ respectively. From cross-validation, $\lambda_c, \lambda_s, \lambda_d$ are set to ($0.2$, $-0.2$, $0.1$), $s_{max} = 0.3$, $m_k=50$, and prototype vectors are randomly initialized. Further details of the experimental settings are provided in Appendix~\ref{apd:setting}.

\vspace{0.1cm}
\noindent\textbf{Evaluation metrics.}  In addition to the area under the receiver operating curve (AUC), we further use global metrics at a low
false alarm rate, similar to \cite{masi2020two}. These metrics require critical detectors to operate at a very low false alarm rate,  especially in the practical use case of automatic screening for fakes on social media. We used the standardized partial AUC or pAUC in addition to the True Acceptane Rate at low False Acceptance Rate. Lastly, we also inspect the Equal Error Rate, similar to \cite{Nguyen2}.

\setlength\extrarowheight{1pt}
\begin{table}[h]
\caption{\textbf{Comparison on FF++ for different quality settings.}
Quantitative results reported for medium compression (c23) and high compression (c40) on FF++ comparing our method with other non-interpretable, interpretable, and temporal methods. Results are reported on four manipulations.} \vspace{0.1cm}
\centering
\label{table:two}
\begin{adjustbox}{width=.97\columnwidth}
\begin{tabular}{c|c|cccc} 
\hline \multirow{2}{*}{Model} & Compression & \multicolumn{4}{c}{FF++} \\ \cline{3-6} 
& Quality   & AUC & $\text{pAUC}_{10\%}$ & $\text{TAR}_{10\%}$ & EER \\ \hline
DSP-FWA \cite{Li} & \multirow{5}{*}{HQ (c23) } & 56.89 & 51.33 & 14.60 & -\\
Xception \cite{Rossler} & & 92.30 & 87.71 & 81.21 & -\\  
Two-branch \cite{masi2020two} & & 98.70 & 97.43 & 97.95 & -\\  \cline{3-6} 
ProtoPNet \cite{Chen} &  & 97.95 & 93.26 & 94.82 & 6.00\\  
DPNet (Ours) &  & \textbf{99.20} & \textbf{98.21} & \textbf{98.04} & 3.41\\\hline

DSP-FWA \cite{Li} & \multirow{5}{*}{LQ (c40) } & 59.15 & 52.04 & 8.82 & -\\
Xception \cite{Rossler} & & 83.93 & 74.78 & 63.25 & -\\  
Two-branch \cite{masi2020two} &  & 86.59 & 69.71 & 62.48 & - \\  \cline{3-6} 
ProtoPNet \cite{Chen} &  & 77.19 & 61.41 & 40.18 & 30.00\\  
DPNet (Ours) &  & \textbf{90.91} & \textbf{81.46} & \textbf{79.46} & 13.35\\\hline
\end{tabular}
\end{adjustbox}
\vspace{-0.6cm}
\end{table}

\setlength\extrarowheight{1pt}
\begin{table*}[ht]
\caption{\textbf{Generalization ability evaluation on unseen datasets.} Our network trained only on FF++ (c23) performs competitively to state-of-the-art baselines while also providing meaningful interpretations. Temporal artifacts learned by our approach can generalize well to datasets with a wider range of visual distortions such as DeeperForensics (Fig. \ref{fig:five}).} \vspace{-0.2cm}
\label{table:three}
\begin{center}
\begin{adjustbox}{width=.95\textwidth}
\begin{tabular}{cc|cccc|cccc|cccc}\hline
\multicolumn{1}{c|}{\multirow{2}{*}{Model}} & \multirow{2}{*}{Train Set} & \multicolumn{12}{c}{Unseen Test Set} \\ \cline{3-14} 
\multicolumn{1}{c|}{} & & \multicolumn{4}{c|}{DFD} & \multicolumn{4}{c|}{DeeperForensics} & \multicolumn{4}{c}{Celeb-DF} \\ \hline
\multicolumn{1}{l}{} & \multicolumn{1}{l|}{} & \multicolumn{1}{l}{AUC} & \multicolumn{1}{l}{$\text{pAUC}_{10\%}$} & $\text{TAR}_{10\%}$ & \multicolumn{1}{l|}{EER} & \multicolumn{1}{l}{AUC} & \multicolumn{1}{l}{$\text{pAUC}_{10\%}$} & $\text{TAR}_{10\%}$ & EER & AUC & $\text{pAUC}_{10\%}$& $\text{TAR}_{10\%}$ & EER \\ \hline
\multicolumn{1}{c|}{Xception \cite{Rossler}} & {\multirow{4}{*}{FF++}}  & 91.27 & 81.25 & \textbf{78.97} & 16.61 & 87.89 & 77.84 & 65.42 & 19.90 & 63.93 & 55.00 & 21.47 & 39.89 \\ 
\multicolumn{1}{c|}{HRNet \cite{wang2020deep}} &  & 89.35 & 79.73 & 71.35 & 18.34 & 83.57 & 74.75 & 62.92 & 22.64 & 61.22 & 54.68 & 19.27 & 45.88 \\ 
\multicolumn{1}{c|}{ProtoPNet \cite{Chen}} & & 84.46 & 77.58 & 66.42 & 23.81 & 61.59 & 52.26 & 14.65 & 38.81 & \textbf{69.33} & \textbf{56.06} & \textbf{29.12} & \textbf{36.52} \\ 
\multicolumn{1}{c|}{DPNet (Ours)} &  & \textbf{92.44} & \textbf{81.30} & 76.21 & \textbf{16.21} &\textbf{ 90.80} &\textbf{ 79.66} & \textbf{75.67} & \textbf{17.30}  & 68.20 & 55.02 & 25.88 & 37.08  \\ \hline
\end{tabular}
\end{adjustbox}
\end{center}
\vspace{-0.6cm}
\end{table*}

\subsection{Evaluations on Different Quality Settings} \vspace{-0.1cm}
In Table \ref{table:two}, we present a thorough comparison of \proposed\ on FF++ \cite{Rossler}, as well as baselines  that do not have interpretable interpretations or temporal aspects. We train and test with four manipulations types (Deepfakes, FaceSwap, Face2Face, and NeuralTextures) along with the real faces for both High Quality (c23) and Low Quality (c40) compressions. We look at different evaluation metrics such as AUC, $\text{pAUC}_{10\%}$, $\text{TAR}_{10\%}$, and EER. In general, our approach has superior performance compared to Xception and Two-branch. In particular, we improved AUC from 92\% to 99\% on HQ (c23) compression, and similarly, AUC from 83\% to 91\% on LQ (c40). The result is also consistent for other low false alarm metrics. Note that Xception and Two-branch does not offer any form of intrinsic interpretability. The table also reports the result of an image-based interpretable method ProtoPNet and a self-supervised method DSP-FWA. Our approach scores the highest AUC across manipulations for both the compression levels, and the usage of temporal information gives {\proposed} an advantage generalizing to more complex and unseen deepfake datasets, seen in the next section.

\subsection{Evaluations on Unseen Datasets} \vspace{-0.1cm}

Table \ref{table:three} shows the benchmark results of our framework on the detection of popular unseen deepfake datasets. We evaluate our model's transferability to Google's DeepfakeDetection, DeeperForensics-1.0, and Celeb-DF, given that it is trained only on FaceForensics++ and \textit{zero} external self-supervised data \cite{li2020face}. For fair comparisons, we trained the model on FF++ with all four manipulation types and real faces, and additionally trained an HRNet-only baseline for binary classification. Table \ref{table:three} reports a clear net improvement over the state-of-the-art, 1-4\% in AUC across the board, even for manipulations with real-world perturbations (DeeperForensics) and low visual artifacts (Celeb-DF). Table \ref{table:four} additionally reports the classic evaluation performance in terms of AUC between our dynamic prototype approach and a wider range of very recent methods on Celeb-DF.

\begin{figure*}[t]
\includegraphics[keepaspectratio=false,height=6cm, width=.98\textwidth]{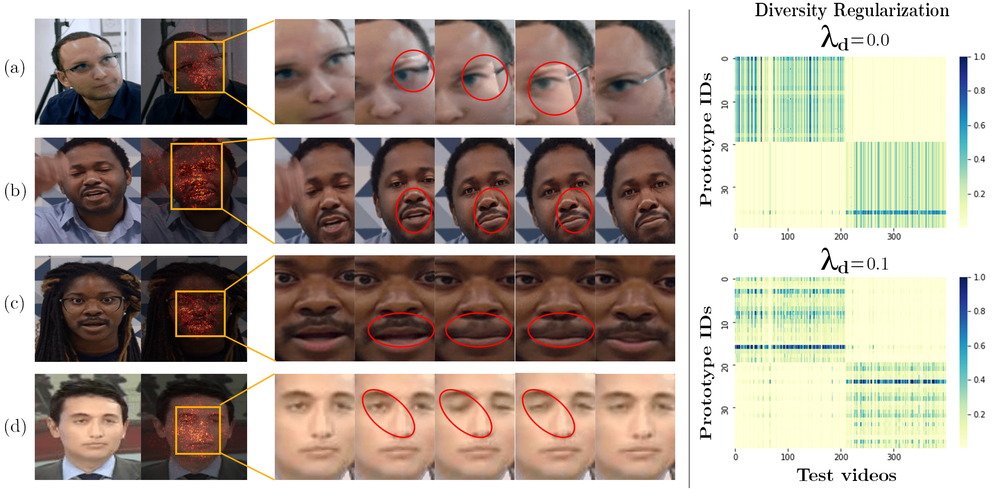}
\caption{\textbf{Visualization of learned dynamic prototypes.} Left column depicts the different classes of temporal artifacts and unnatural movements found by {\proposed}. Right column shows the effect of diversity regularization~\eqref{eq:div} on prototype similarities across test videos.}
\label{fig:four}
\vspace{-0.3cm}
\end{figure*}

\noindent \textbf{The effect of temporal information.} Our network makes predictions about face forgery by exploiting temporal artifacts, which are harder to hide and can generalize across different manipulations in the RGB pixel space. Table \ref{table:two} indicates general improvement over not only Xception, but also the image-based interpretable ProtoPNet, where the gap is much wider as visual quality degrades. Patch-based comparison in $\ell_2$-space looks at the visual patch (enforced by the loss), and significant changes to the video visuals can impact interpretability. This can be further seen in Table \ref{table:three}, where both ProtoPNet and DPNet performs comparably on Celeb-DF, but differ significantly for DeeprForensics. This is not surprising as DeeperForensics features a wider range of distortions, using compressions, color saturations, contrast changes, white gaussian noises, and local block-wise distortions \cite{jiang2020deeperforensics10} (Figure \ref{fig:five}). And since the number of prototypes are fixed, high performance requires generalizable prototypes. Table \ref{table:five} further investigates the impact of varying the number of flow frames.
\vspace{0.1cm}

\noindent \textbf{Qualitative visualization of learned  dynamic prototypes.}
Figure \ref{fig:four} presents classes of temporal artifacts captured by the learned dynamic prototypes. We take the gradient attribution with respect to each prototype similarity score. During the training steps, while projecting, we kept track of the latent patch within the original training clip that is closest to each prototype. Hence, during testing, we can visualize the prototypes that are most similar to the testing videos. Some artifacts features heavy discoloration (Fig. \ref{fig:four}a), which is already interpretable as images, but changes in discoloration over time offer more interpretability. In fact, other artifacts such as subtle discolorations or movements (Fig. \ref{fig:four}b,c) are much harder to interpret with just one image, especially the facial jitterings and unnatural oscillations (Fig. \ref{fig:four}d).

\begin{table}[ht]
\begin{minipage}[t]{0.6\columnwidth}
\caption{\textbf{Best competing methods on Celeb-DF are reported}. Results for other methods are from \cite{li2020celeb}} \vspace{0.1cm}
\begin{adjustbox}{width=\columnwidth}
\begin{tabular}{l|cc} \hline
\multirow{1}{*}{Model}  & \multirow{1}{*}{FF++} & \multirow{1}{*}{Celeb-DF}\\ \hline
MesoInception4 \cite{Afchar} & 83.0 & 53.6 \\
Two-stream \cite{Zhou} & 70.1 & 53.8 \\
HeadPose \cite{Yang} & 47.3 & 54.6 \\
VA-MLP \cite{Matern} & 66.4 & 55.0 \\
VA-LogReg & 78.0 & 55.1 \\
Xception-raw \cite{Rossler} & \textbf{99.7} & 48.2 \\
Xception-c23 & \textbf{99.7} & 65.3 \\
Xception-c40 & 95.5 & 65.5 \\
Multi-task \cite{Nguyen2} & 76.3 & 54.3  \\
Capsule \cite{Nguyen1} & 96.6 & 57.5 \\
DSP-FWA \cite{Li} & 93.0 & 64.6 \\ \hline
ProtoPNet \cite{Chen} & 98.0 & 69.3 \\
DPNet (Ours) - c23 & 99.2 & 68.2 \\
DPNet (Ours) - c40 & 90.91 & \textbf{71.76} \\ \hline
\end{tabular}
\label{table:four}
\end{adjustbox}
\end{minipage}\hfill
\begin{minipage}[t]{0.37\columnwidth}
\captionof{figure}{\textbf{Visual distortions in DeeperForensics} \cite{jiang2020deeperforensics10}. Not in FF++.} \vspace{0.1cm}
\begin{adjustbox}{width=\columnwidth}
\includegraphics[keepaspectratio=false,height=18cm, width=10cm]{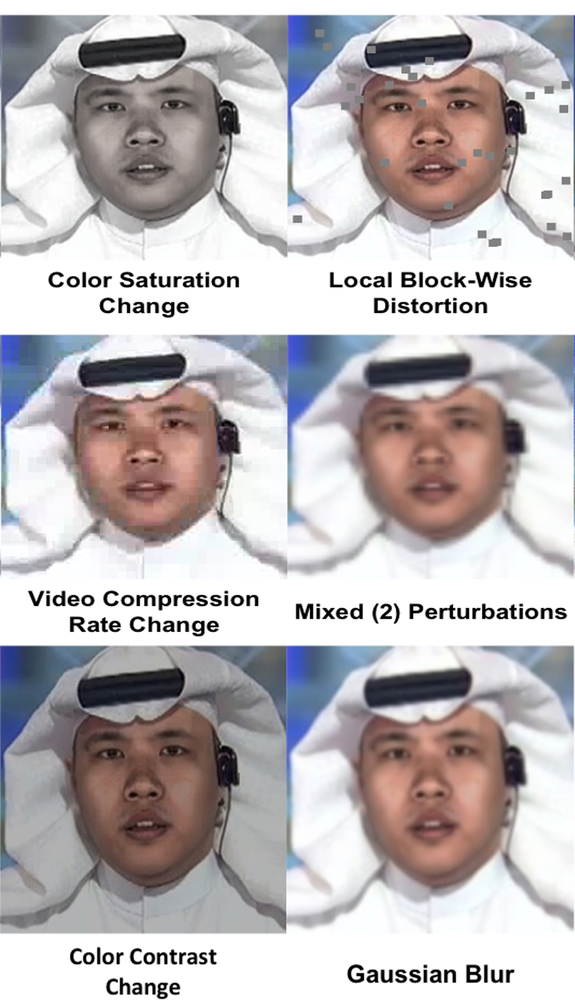}
\label{fig:five}
\end{adjustbox}
\end{minipage}
\vspace{-0.6cm}
\end{table}

\begin{table*}
    \centering
    \caption{\textbf{Ablation study on FF++.} (a) Testing metrics obtained by training varying the number of prototypes $m_k$ under (c23) compression, testing on unseen DFD. (b) Ablation experiments showing the impact of length of the input flow segments, testing on unseen DeeperForensics (c) Ablation experiments on (c23) compression without diversity regularization.} \vspace{0.2cm}
    \begin{adjustbox}{width=.3\textwidth}
    \begin{tabular}{c|cccc}\hline 
    \multirow{2}{*}{(a)} & \multicolumn{4}{c}{FF++ - DFD - varying num prototypes} \\ \cline{2-5} 
     & AUC & $\text{pAUC}_{10\%}$ & $\text{TAR}_{10\%}$ & EER \\ \hline
    $m_k$ = 10 & 86.19 & 74.09 & 56.13 & 21.93\\
    $m_k$ = 25 & 90.79 & 77.36 & 71.38 & 17.46\\
    $m_k$ = 50 & 92.44 & 81.30 & 76.21 & 16.21\\
    $m_k$ = 100 & 91.88 & 79.94 & 77.70 & 17.1\\ \hline
    \end{tabular}
    \end{adjustbox}
    \hfill
    \begin{adjustbox}{width=.375\textwidth}
    \begin{tabular}{c|cccc}\hline 
    \multirow{2}{*}{(b)} & \multicolumn{4}{c}{FF++ - DeeperForensics - wo/ flow frames} \\ \cline{2-5} 
     & AUC & $\text{pAUC}_{10\%}$ & $\text{TAR}_{10\%}$ & Time per Batch \\ \hline
    - \texttt{flows}  & 61.59 & 52.26 & 14.65 & 0.68 $\pm$ 0.13s\\
    \texttt{frame} = 5 & 88.03 & 77.76 & 69.38 & 1.11 $\pm$ 0.13s\\
    \texttt{frame} = 10 & 90.80 & 79.66 & 75.67 & 2.22 $\pm$ 0.44s\\
    \texttt{frame} = 15 & 84.85 & 75.22 & 66.98  & 4.39 $\pm$ 0.34s\\ \hline
    \end{tabular}
    \end{adjustbox}
    \hfill
    \begin{adjustbox}{width=.3\textwidth}
    \begin{tabular}{c|ccc}\hline 
    \multirow{2}{*}{(c)} & \multicolumn{3}{c}{FF++ c23 - wo/ Diversity} \\ \cline{2-4} 
     & AUC & $\text{pAUC}_{10\%}$ & $\text{TAR}_{10\%}$ \\ \hline
    - \texttt{diversity} & 97.81 & 96.17 & 95.76\\
    DPNet & 99.20 & 98.21 & 98.04\\ \hline
    \end{tabular}
    \end{adjustbox}
    \label{table:five}
    \vspace{-0.4cm}
\end{table*}

\subsection{Temporal specifications over prototypes} \vspace{-0.1cm}
\begin{table}[h]
\caption{\textbf{Percentage of traces satisfying temporal specifications.} Rows in each block represent the percentage over positive, negative, and all traces.} 
\vspace{0.1cm}
\centering
\begin{adjustbox}{width=.85\columnwidth}
\begin{tabular}{ c|c|ccc }
\hline
&  & $\phi_{1,key\_frame}$ & $\phi_{2, non\_relevance}$ & $\phi_{3, relaxed}$ \\
\hline
\multirow{3}{4em}{ProtoPNet} & (+) & 95.23 & 49.73 & 70.89 \\ 
& (-) & 89.09 & 42.18 & 58.76 \\ \cline{2-5}
& & 92.00 & 45.75 & 64.50 \\ \hline
\multirow{3}{4em}{DPNet (Ours)} & (+) & 93.65 & 28.04 & 74.07 \\ 
& (-) & 91.46 & 16.11 & 56.39 \\ \cline{2-5}
& & 92.50 & 21.75 & 64.75 \\ \hline
\end{tabular}
\end{adjustbox}
\label{table:six}
\vspace{-0.2cm}
\end{table}

Table \ref{table:six} reports the robustness of {\proposed} and interpretable baselines against desired temporal specifications specified in Section \ref{ss:temporal}. Overall, both approaches satisfy the $\phi_{1,key\_frame}$ specification up to high-percentage, with {\proposed} performing better, especially with the $fake$ traces. This indicates that with high-probability, we can find a key-frame within the video that is most relevant in explaining why a video is a deepfake, via a dynamic prototypes in the $fake$ class. On the other hand, the models perform poorly with the stringent specification $\phi_{2, non\_relevance}$, which requires non-relevant prototypes to both stay low and not change at all over time. Since {\proposed} utilized temporal information via optical flows, this is a stricter specification to enforce as flows can change drastically across consecutive frames, hence the lower percentage of satisfying traces. We experiment by relaxing the consecutive ``next 5 frames" non-changing constraint, $\phi_{3, relaxed}$, which now only enforce non-relevant prototype similarities to be low. The flexibility of temporal logic along with interpretable model allows end-users to specify and enforce certain desiderata in their detection framework. This further increases trustworthiness and interpretability of these detection frameworks.

\subsection{Ablation study} \vspace{-0.1cm}

\noindent\textbf{Choosing the number of prototypes $m_k$.} Table \ref{table:five}a shows the ablation experiments when choosing the number of prototypes. We further investigate the generalization impact of this decision on unseen deepfakes (DFD). Using the same
hyperparameter configuration, the parameter is set $m_k = \{10,25, 50, 100\}$. As shown in Table \ref{table:five}a , the AUC first improves dramatically as $m_k$ increases, but continuing to do so yields diminishing returns. Thus, there is an inherent trade-off between predictive performance and interpretability. In practice, since increasing k after a certain threshold only brings marginal improvement to the performance, we would suggest to gradually increase $m_k$, until a desired low false alarm threshold is achieved.

\noindent\textbf{Ablation study on the number of flow frames.} Table \ref{table:five}b further reports the generalization performance to DeeperForensics when varying the number of flow frames utilized by the network. Here we test the generalization of the dynamic prototypes and how the length of the time segment given to the network impacts its performances. As shown in Figure \ref{fig:five}, visual elements of deepfakes can change drastically across domains and manipulations. However, temporal information can be key in pinpointing artifacts shared by all manipulations. Using the same configuration as previous, the number of flow frames is set $\texttt{frame} = \{0, 5, 10, 15\}$, where zero frames do not use flows, similar to the ProtoPNet baseline. In general, Table \ref{table:five}b reports a significant net improvement on the unseen dataset DeeperForensics, where there are a larger number of visual distortions. In addition, we observe that using a small number of flow frame can already yield a better performance, and that using a longer sequence can negatively impact prediction as the inputs become noisier. Moreover, the trade-off between predictive performance and training time will also have to be considered since using more frames significantly impact each batch loading and processing time.

\begin{figure}[t]
\centering
\includegraphics[width=.8\columnwidth]{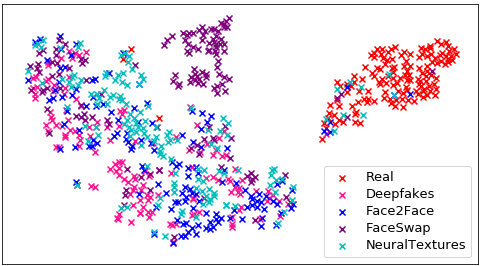}
\caption{\textbf{The t-SNE embedding visualization of the DPNet trained for binary classification on FF++ (c23)}. Red X's are real videos, others represent data generated by different manipulations.}
\label{figure:six}
\vspace{-0.5cm} 
\end{figure}
 
 \noindent\textbf{The effect of diversity regularization $\lambda_d$.} To study the
effect of the diversity regularization term, we removed the term by setting $\lambda_d = 0$ measure the AUC drop on FF++. Results are shown in Table \ref{table:six}c. We further examine the impact of $\lambda_d$ by plotting the similarity scores between the prototypes and test videos as heatmaps (Figure \ref{fig:four}-right). Without the diversity regularization, most of the rows have similar patterns, indicating that the prototypes are duplicating and are less diverse, yielding lower test AUC on FF++.
\vspace{-0.3cm}

\section{Conclusion} \vspace{-0.2cm}
There is a growing need to use automatic deepfake detection models to detect and combat deepfakes. However, a reliance on deepfake detectors places enormous trust on these models. This work aims to help justify this trust by improving the interpretability of deepfake detection. In addition to model interpretability, this work offers insights into what parts of deepfake videos can be used to discern deepfakes, which may inform people how to detect deepfakes themselves. Model interpretations strengthen the accountability of deepfake detectors and our work encourages future research in explaining deepfake detectors. 

\vspace{0.2cm}
\noindent \textbf{Acknowledgment}

\noindent \small This work is supported by the National Science Foundation under Grant No. 1619458, the FMitF program, Grant No. CCF-1837131, and the Defense Advanced Research Projects Agency (DARPA) under Agreement No. HR00111990059.

{\small
\bibliographystyle{ieee_fullname}
\bibliography{egbib}
}

\clearpage
\begin{appendix}
\section*{Appendix}

\section{Experimental Settings} \vspace{-0.1cm}\label{apd:setting}
\textbf{Datasets.} The FaceForensics++ (FF++) dataset \cite{Rossler} consists of 1000 original video sequences sourced from \texttt{YouTube} and 1000 synthetic videos generated using four different manipulation techniques. The Google/Jigsaw DeepFake detection (DFD) \cite{DDD_GoogleJigSaw2019} dataset consists of 3,068 deepfake videos generated based on 363 original videos of 28 consented individuals of various ages, genders, and ethnic groups. Pairs of actors were selected randomly and deep neural networks swapped the face of one actor onto the head of another. The specific deep neural network synthesis model is not disclosed, but manipulation masks are also provided in this dataset. DeeperForensics-1.0 \cite{jiang2020deeperforensics10} is a large scale dataset consisting of 11, 000 deepfake videos generated with high-quality collected data vary in identities, poses, expressions, emotions, lighting conditions, and 3DMM blendshapes. Real-world distortions such as compressions, color saturations, contrast changes, white gaussian noises, and local block-wise distortions, are heavily applied and mixed.  Celeb-DF \cite{li2020celeb} is a new DeepFake dataset of celebrities with 408 real videos and 795 synthesized video with reduced visual artifacts (Table \ref{table:one}).

\textbf{Pre-processing.} We use a standard data pre-processing technique for deepfakes and videos, i.e. dumping the video frames to images and cropping out facial areas using facial boundaries and landmarks \cite{mtcnn} instead of using the full-frame. When cropping faces, we use a face margin of 0.3-0.5 to get a full cropping of the actors' head. Consecutive are processed in OpenCV to calculate the dense optical flows and cropped similarly to the RGB frames.

\textbf{Training details.}
During training, each input is formed by 1 RGB frame with 10 pre-computed optical flow fields starting from that RGB frame. Given the FPS of FF++, this registers a temporal signature of roughly 0.5s. We sample 270 frames from each FF++ training video \cite{Rossler} and utilized the standard normalization on the RGB frame, but not the flow frames. Besides this, we do not use any other form of video data augmentations. We used a pre-trained HRNet as the backbone of our architecture. However, given the shape of our input, we also need to perform \textit{cross modality pre-training} to initialize the weights of the first conv. layer by averaging the weights across the RGB channels and replicate it corresponding to the number of input flow channels \cite{Wang}. Our network parameters $\theta$ are trained using Adam, with a learning rate of $2e^{-4}$ for $f_\omega$ and $1e^{-3}$ for $\textbf{p}_i$.

\textbf{Validation details.}
During validation, we sample 100 frames from each testing video, and do the same across all unseen datasets. We combine the logits using an aggregation function (\texttt{sum} or \texttt{avg}). This can be understood as combining the \textit{similarity evidence} between the learned prototypes and the testing video across clips in the video.

\section{TQTL Semantics} \vspace{-0.1cm} \label{apd:TQTL}
For completeness, we detailed the semantics for TQTL for evaluating a specification, similar to the work by Dokhanchi et al. \cite{Dokhanchi}. Consider the data stream $\mathcal{D}, i \in N$ is the
index of current frame, $\pi \in P$, $\phi, \phi_1, \phi_2 \in TQTL$ and evaluation function $\epsilon: V_t \cup V_p \rightarrow \mathbb{N}$, which is the environment over the time and prototype variables. The quality value of formula $\phi$ with respect to $\mathcal{D}$ at frame i with
evaluation $\epsilon$ is recursively assigned as follows:
\begin{align*}
\llbracket \top \rrbracket (\mathcal{D}, i, \epsilon) &\coloneqq +\infty\\
\llbracket\pi \rrbracket(\mathcal{D}, i, \epsilon) & \coloneqq \llbracket f_\pi (t_{1\dots n}, p_{1\dots n})\sim c\rrbracket (\mathcal{D}, i , \epsilon)  \\
\llbracket x.\phi \rrbracket (\mathcal{D}, i, \epsilon) &\coloneqq \llbracket \phi \rrbracket (\mathcal{D}, i , \epsilon[x \Leftarrow i])\\
\llbracket \exists p_i @ x, \phi \rrbracket (\mathcal{D}, i, \epsilon) &\coloneqq \max_{k \in \mathcal{P}}{(\llbracket\phi\rrbracket(\mathcal{D}, i, \epsilon[p_i \Leftarrow k ] )) }\\
\llbracket x\leq y  + n \rrbracket (\mathcal{D}, i, \epsilon ) & \coloneqq \left\{\begin{array}{lr} + \infty \quad if  ~\epsilon(x) \leq \epsilon(y) + n \\
-\infty\quad otherwise
    \end{array}\right.\\
\llbracket \neg\phi \rrbracket(\mathcal{D}, i , \epsilon) &\coloneqq -  \llbracket \phi \rrbracket (\mathcal{D}, i, \epsilon) \\
\llbracket \phi_1 \vee \phi_2\rrbracket(\mathcal{D} ,i, \epsilon) &\coloneqq \max\left(\llbracket \phi_1 \rrbracket(\mathcal{D}, i, \epsilon), \llbracket\phi_2\rrbracket(\mathcal{D}, i, \epsilon) \right)  \\
\llbracket \phi_1 U \phi_2 \rrbracket(\mathcal{D}, i, \epsilon) &\coloneqq \max_{i \leq j }{\Big(\min \big( \llbracket \phi_2 \rrbracket (\mathcal{D}, j, \epsilon),} \\
&\hspace{1.5cm}{\min_{i\leq k < j} \llbracket \phi_1 \rrbracket (\mathcal{D}, k, \epsilon \big) \Big)}
\end{align*}
\noindent We say that $D$ satisfies $\phi$ $(\mathcal{D} \models \phi)$ iff $\llbracket \phi \rrbracket (\mathcal{D}, 0, \epsilon_0) > 0$, 
where $\epsilon_o$ is the initial environment. On the other hand, a data stream $\mathcal{D}'$ does not satisfy a TQTL formula $\phi$ $(\mathcal{D}' \not\models \phi)$, iff $\llbracket \phi \rrbracket (\mathcal{D}, 0, \epsilon_0) \leq 0$. The quantifier $\exists id@x$ is the maximum operation on the quality values of formula $\llbracket \phi \rrbracket$
corresponding to the prototypes IDs at frame $x$.

\section{GIFs} \vspace{-0.1cm} \label{apd:gifs}
For the following visualizations and figures in the main paper, we have included the corresponding gifs in folders within the supplementary materials.

\section{Additional Visualizations} \vspace{-0.1cm}
\subsection{More examples of prototypes and classes of temporal artifacts learned.}
\subsection{More examples of how {\proposed} classify a video.}

\begin{figure*}[h]
\includegraphics[keepaspectratio=false,height=9cm, width=\textwidth]{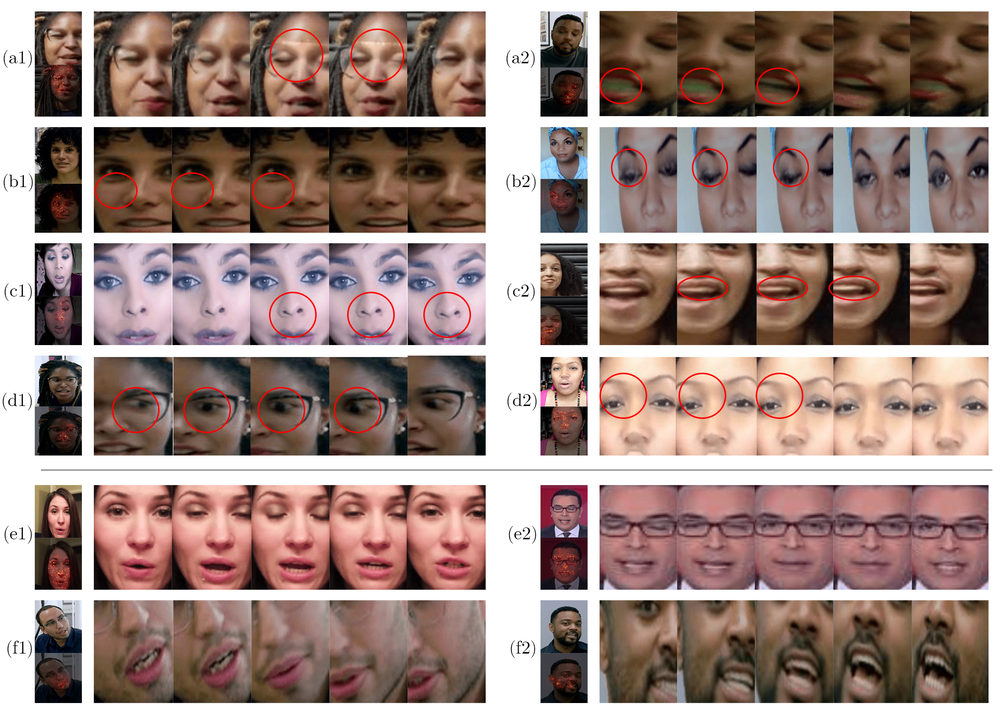}
\caption{\textbf{Different classes of temporal artifacts and unnatural movements found by {\proposed}}. Top block are fake prototypes and bottom are real prototypes. a) heavy discolouration, b) subtle discolouration, c) subtle disappearance, d) unnatural movement, e) combined eye-mouth movement, and f) head movement. Best view as GIFs.}
\end{figure*}

\begin{figure*}[h]
\centering
\includegraphics[keepaspectratio=false,height=20cm, width=0.85\textwidth]{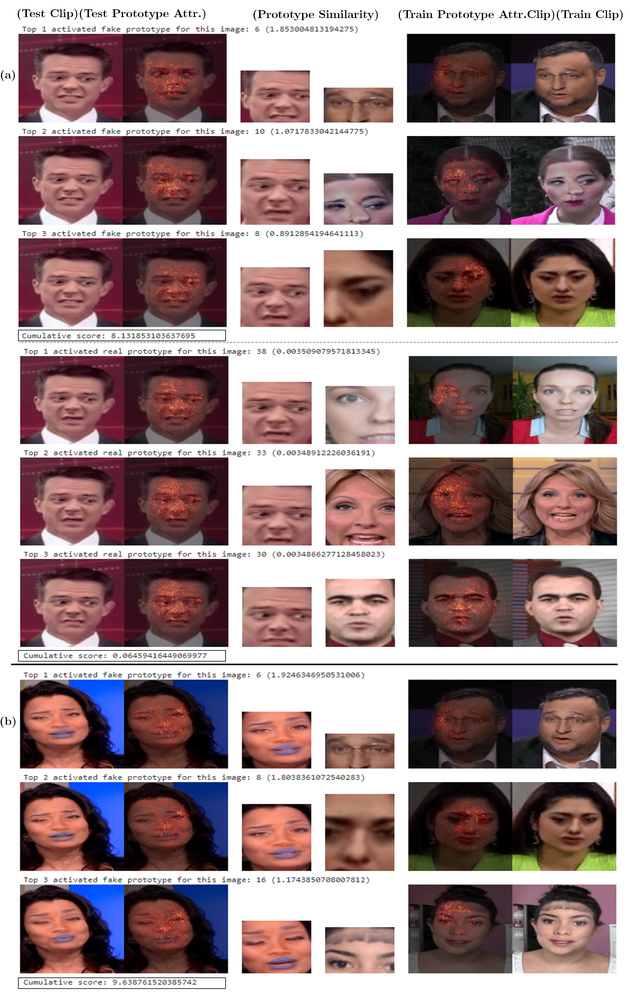}
\end{figure*}
\newpage

\begin{figure*}[h]
\centering
\includegraphics[keepaspectratio=false,height=20cm, width=0.85\textwidth]{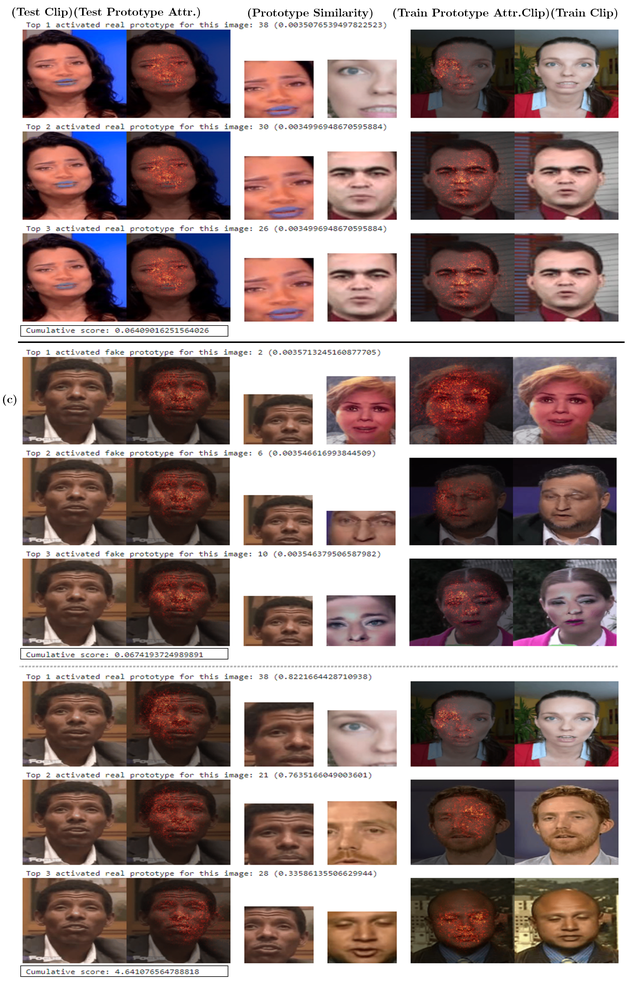}
\caption{\textbf{More examples of how {\proposed} classify a video.} (a) and (b) are deepfakes, and (c) is genuine. Best view as GIFs. The prediction for each class is based on the evidence between the dynamics of the input and a small set of dynamic prototypes. }
\end{figure*}

\end{appendix}

\end{document}